%% file: fbv.tex
\documentclass[%
                10pt,%
                a5paper,%
                headings=small,%
                listof=totoc,%
                headsepline,%
                DIV=calc,%
                BCOR=12mm,%
                captions=tableheading,%
                fleqn,%
                runningheads,%
                ]{fbvproc}

\usepackage{etex}
\usepackage{makeidx}
\usepackage[ngerman,english]{babel}
\usepackage{graphicx}
\usepackage[utf8]{inputenc}
\usepackage{tabularx}
\usepackage{amsfonts}
\usepackage[mathscr]{eucal}
\usepackage{listings}
\usepackage[small,bf,normal]{caption}
\usepackage{color}
\usepackage[table]{xcolor}
\usepackage[standard,thmmarks]{ntheorem}
\usepackage{url}
\usepackage[amssymb]{SIunits}
\usepackage{bibunits}
\usepackage{amsmath}
\usepackage{subfigure}
\usepackage{cite}
\usepackage{nicefrac}
\usepackage{palatino}

\usepackage{dsfont}
\usepackage{enumerate}
\usepackage{wrapfig}
\usepackage{tikz}
\usetikzlibrary{matrix,arrows} 
\usepackage{wasysym}
\usepackage[babel,german=quotes]{csquotes}

\usepackage{xspace}

\usepackage{hyperref}
\usepackage{chngcntr}
 
\counterwithout{figure}{chapter}
\counterwithout{table}{chapter}

\newenvironment{itemize*}%
  {\begin{itemize}%
    \setlength{\itemsep}{1pt}%
    \setlength{\parskip}{1pt}}%
  {\end{itemize}}

\newenvironment{enumerate*}%
  {\begin{enumerate}%
    \setlength{\itemsep}{1pt}%
    \setlength{\parskip}{1pt}}%
  {\end{enumerate}}

\newenvironment{description*}%
  {\begin{description}%
    \setlength{\itemsep}{1pt}%
    \setlength{\parskip}{1pt}}%
  {\end{description}}

\begin{document}
\selectlanguage{ngerman}
\thispagestyle{empty}

\newpage
\thispagestyle{empty}
\cleardoublepage
\frontmatter          


\mainmatter              



\input{schauwecker.tex}          

\renewcommand{\bibname}{Literatur}

\value{chapter}=9
\addcontentsline{toc}{chapter}{Index}

\end{document}

%% file: schauwecker.tex

\selectlanguage{english}
\hyphenation{Scene-Scan}

\title{Real-Time Stereo Vision on FPGAs with SceneScan}

%
\titlerunning{FPGA Stereo Vision}  
%
\author{Konstantin Schauwecker\inst{1}}
\authorrunning{K. Schauwecker}   
%
\tocauthor{K. Schauwecker}
\institute{Nerian Vision GmbH,\\
Gutenbergstr. 77a, 70197 Stuttgart, Germany\\
\url{www.nerian.com}}

\abstract{
We present a flexible FPGA stereo vision implementation that is capable of processing up to 100 frames per second or image resolutions up to 3.4 megapixels, while consuming only 8~W of power. The implementation uses a variation of the Semi-Global Matching (SGM) algorithm, which provides superior results compared to many simpler approaches. The stereo matching results are improved significantly through a post-processing chain that operates on the computed cost cube and the disparity map. With this implementation we have created two stand-alone hardware systems for stereo vision, called SceneScan and SceneScan Pro. Both systems have been developed to market maturity and are available from Nerian Vision GmbH.
}

\maketitle              

\section{Introduction}

Computer stereo vision is one of the best researched areas in the field of computer vision. Its origins date back to the 1970s, and it has since seen significant scientific advancement. Compared to other approaches for dense depth perception -- such as time-of-flight cameras or structured light camera systems -- stereo vision has the advantage of being a passive technology. This means that apart from the ambient light, no further active light source is required for performing measurements.

Active camera systems struggle in situations with bright ambient light such as outdoors during bright sunlight. In this case, the active light source can no longer provide sufficient contrast from the ambient light, and hence measurements become impossible. Likewise, long-range measurements are challenging for active camera systems as this would require a particularly powerful light source. Stereo vision is consequently a promising technology for outdoor and long-range use cases, and can enable depth sensing for applications where other technologies are not applicable.

For autonomous mobile robots that need to operate outside of a controlled indoor environment, robustness towards the lighting situation is critical. Fast-moving robots, such as autonomous vehicles, must also be able to perform long-range measurements in order to operate safely. It are thus these applications that require a robust sensor solutions like real-time stereo vision.

Despite this need for robust depth sensing, there are currently only few applications that make use of stereo vision. The key problem that has prevented a wider spread adoption of this technology is the high computational demand. While many algorithms for stereo analysis have been proposed in literature within the past decades, good-performing methods are still very computationally demanding, even on modern hardware.

Traditional CPUs struggle with the vast number of computations that are required for stereo vision at typical camera frame rates. For simpler algorithms -- such as block matching -- real-time stereo vision can be achieved on a CPU, as has for example been demonstrated in \cite{HUM2010}. Such simple algorithmic approaches significantly lack behind more modern methods in terms of accuracy and error robustness, as can be seen in the KITTI vision bechnmark \cite{GEIG2018}. There has been some success creating optimized implementations of more sophisticated methods \cite{GEH2010,SPAN2014}, but the overall computational burden remains high and the achieved processing rates and image resolutions are still comparatively low.

In order to speed-up the image processing, one requires hardware that is capable of massively parallel computations. One candidate hardware platform are graphics processors or GPUs. Numerous works in this field have demonstrated that fast real-time stereo vision is possible on GPUs with optimized algorithm implementations \cite{HAL2010,KOW2013}. These works focused primarily on using high-end GPUs with outstanding computational capabilities. Unfortunately, such high-end GPUs also exhibit a very high-power consumption. The GeForce GTX TITAN X by NVIDIA for example, requires up to 250 W of power \cite{NVID2018}. Deploying such a power hungry processing hardware on a battery-powered mobile system is often impractical, which is why GPU-based stereo vision has not yet seen a wide spread use in mobile robotics.

An alternative hardware platform to GPUs are Field Programmable Gate Arrays (FPGAs). An FPGA is a generic integrated circuit that can be programmed to fulfill a particular application. Because programming is performed on the circuit level, an FPGA is not forced to follow the usual fetch-decode-execute cycle of CPUs or even modern GPUs. Rather, an application specific architecture can be found that divides the problem into many small sub-problems, which can each be solved in parallel and with minimal power consumption.

The disadvantage of FPGAs is that the programming effort vastly exceeds the effort of programming ordinary CPUs or GPUs. Thus, FPGA-based image processing is not yet commonplace in today's vision systems. In order to make this technology more accessible to researches and application developers, Nerian Vision GmbH\footnote{\url{http://nerian.com}} developed the SceneScan and SceneScan Pro stereo vision sensor systems. The more capable SceneScan Pro system is shown in Figure~\ref{fig:scenescan}. Using an FPGA, this small-scale processing device is able to perform stereo vision in real-time and at high processing rates. Its key features are:

\begin{figure}[tbp]
    \centering
    \includegraphics[width=\textwidth,trim={0 5cm 0 5cm},clip]{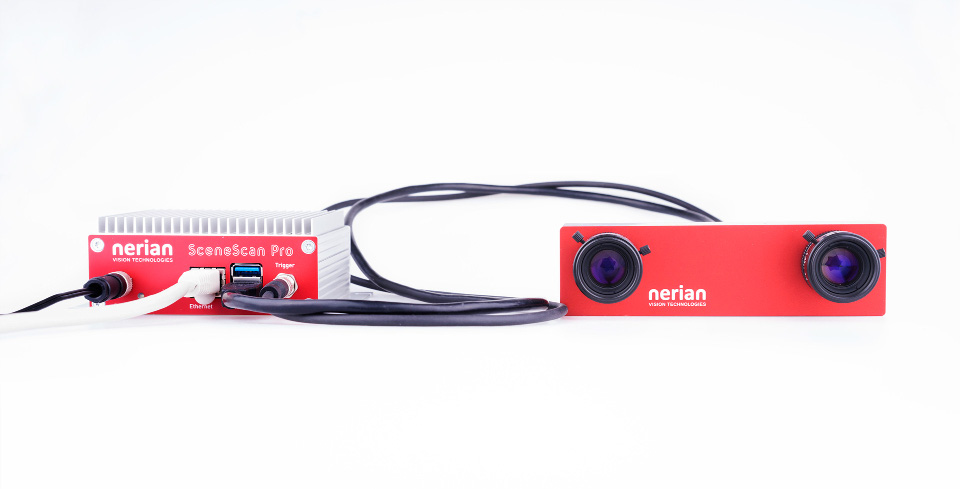}
    \caption {SceneScan Pro with connected Karmin2 stereo camera.}
    \label{fig:scenescan}
\end{figure}

\begin{itemize}
\item Processes input images with resolution up to 3.4 megapixels.
\item Covers a disparity range of up to 256 pixels, with a sub-pixel resolution of $\nicefrac{1}{16}$ pixel.
\item Can process up to 100 frames per second.
\item Power consumption of 8 W.
\item Computed disparity map is transmitted in real-time through gigabit ethernet.
\end{itemize}

SceneScan and SceneScan Pro are the successors to Nerian's earlier SP1 system \cite{SCH2015}. Compared to the SP1, SceneScan Pro has tripled the processing performance and improved the measurement quality significantly. This article provides a review of how this has been possible and gives an insight into the achieved performance.

\section{Architecture}

SceneScan / SceneScan Pro connect to Nerian's own Karmin2 stereo camera or to two standard USB3 Vision cameras. It acquires a pair of synchronized left and right camera images and then processes these images on its built-in FPGA. The computed stereo disparity map, which is an inverse depth map, is then output through gigabit ethernet to a connected computer.

The stereo matching method that is implemented in SceneScan / SceneScan Pro is based on Semi Global Matching (SGM) \cite{HIR2005}. Since its proposal, SGM has enjoyed much popularity, due to its high-quality results and its computational efficiency. SGM is at the heart of some of the best performing stereo matching methods, such as the one proposed by \v{Z}bontar et al. \cite{ZBO2015}.

\begin{figure}[tbp]
    \includegraphics[width=\textwidth]{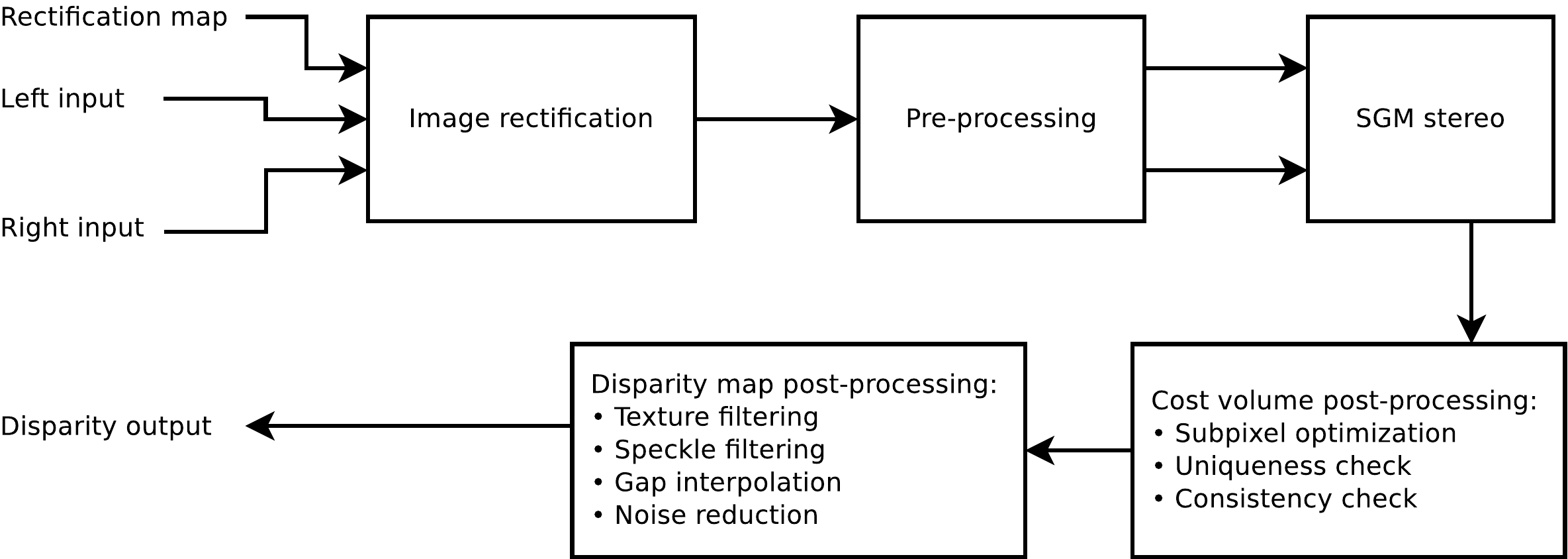}
    \caption {Block diagram of processing structure.}
    \label{fig:overview}
\end{figure}

It has been shown that FPGA-based implementations of SGM are possible \cite{GEH2009,BAN2010}. However, only relying on SGM is not sufficient for receiving competitive stereo matching results. Rather, we require an entire image processing pipeline, which not only includes SGM, but also a range of different pre- and post-processing steps. We have implemented one such pipeline for SceneScan / SceneScan Pro, which is depicted in Figure~\ref{fig:overview}.

\subsection{Rectification}
\label{sec:rectification}

The first processing step that is applied to the input image data is image rectification. To perform image rectification, a pre-computed rectification map is required that is read from a dedicated memory. The rectification map contains a subpixel accurate x- and y-offset for each pixel of the left and right input images. Bi-linear interpolation is applied to map the subpixel offsets to image intensities.

The offsets are interleaved such that reading from a single data stream is sufficient for finding the displacement vector for each pixel in both images. To save bandwidth, the rectification map is stored in a compressed form. On average, one byte is required for encoding the displacement vector for a single pixel. Hence, the overall size of the rectification map is equal to the size of two input images.

Rectification is a window-based operation. Hence, the pixel offsets are limited by the employed window size. For SceneScan and SceneScan Pro a window size of $79 \times 79$ pixels is used. This allows for offsets in the range of $-39$ to $+39$ pixels.

\subsection{Image Pre-Processing}
\label{sec:census}

An image pre-processing method is applied to both input images. This causes the subsequent processing steps to be more robust towards illumination variations and occlusions.

\subsection{Stereo Matching}
\label{sec:sgm}

Stereo matching is performed by applying a variation of the SGM algorithm \cite{HIR2005}. The penalties $P_1$ and $P_2$ that are employed by SGM for small and large disparity variations can be configured at runtime. Several iterations are required for processing one pixel of the left input image. In each iteration, the left image pixel is compared to a group of pixels in the right image. For SceneScan Pro, the number of parallel pixel comparisons is $p = 32$. The number of iterations per left image pixel $n_i$ can be configured at runtime by choosing the disparity range. 

A \textit{disparity offset} $o_d$ can also be configured at runtime, which indicates the smallest disparity value that will be considered during stereo matching. If $o_d > 0$ then the observable depth range will have an upper limit, as disparities smaller than $o_d$ will not be allowed. The disparity offset $o_d$, iteration count $n_i$ and the parallelization $p$ determine the maximum disparity $d_{max}$:
\begin{equation}
    d_{max} = o_d + n_i p - 1
\end{equation}

\subsection{Cost Volume Post-Processing}

The SGM stereo algorithm produces a \textit{cost volume}, which encodes the matching costs for all valid combinations of left and right image pixels. Several of the applied post-processing techniques operate directly on this cost volume.

\subsubsection{\small Subpixel Optimization}

\label{sec:subpixel}
Subpixel optimization is the first applied post-processing technique. This step increases the accuracy of depth measurements by evaluating the matching costs to the left and right of the detected minimum for each pixel. A curve is fitted to the matching costs and its minimum is determined with subpixel accuracy.

The improved disparity estimates are then encoded as fixed-point numbers. SceneScan / SceneScan Pro support 4 decimal bits for the subpixel optimized disparity. Hence, it is possible to measure disparities with a resolution of $\nicefrac{1}{16}$ pixel.

\subsubsection{\small Uniqueness Check}
\label{sec:uniq}

Matches with a high matching uncertainty are discard by imposing a uniqueness constraint. For a stereo match to be considered unique, the minimum matching cost $c_{min}$ times a uniqueness factor $q \in [1, \infty)$ must be smaller than the cost for the next best match. This relation can be expressed in the following formula, where $C$ is the set of matching costs for all valid pixel pairs and $c^* = c_{min}$ is the cost for the best match:
\begin{equation}
\label{eqn:unique}
c^* \cdot q < \min \left\{C \setminus \{c_{\min} \}\right\} \mbox{.}
\end{equation}

Stereo matches that are discarded through the uniqueness check are assigned an invalid disparity label.

\subsubsection{\small Consistency Check}
\label{sec:consist}

A consistency check is employed for removing further matches with high matching uncertainties. The common approach to this post-processing technique is to repeat stereo matching in the opposite matching direction (in our case from the right image to the left image), and then only retaining matches for which
\begin{equation}
    | d_l - d_r | \le t_c \mbox{,}
\end{equation}
where $d_l$ is the disparity from left-to-right matching, $d_r$ the disparity from right-to-left matching, and $t_c$ is the consistency check threshold.

In order to save FPGA resources, we refrain from re-running stereo matching a second time in the opposite matching direction. Rather, the right camera disparity map is inferred from the matching costs that have been gathered during the initial left-to-right stereo matching. Pixels that do not pass the consistency check again receive an invalid disparity label.

\subsection[Disparity Map Post-Proc.]{Disparity Map Post-Processing}

Following the cost volume post-processing, the cost volume is reduced to a disparity map. Additional post-processing methods are then applied directly to the disparity values.

\subsubsection{\small Texture Filtering}
\label{sec:texture_filt}

Matching image regions with little to no texture is particularly challenging. Especially if such regions occur close to image borders, this might lead to significant mismatches. In order to address this problem, a texture filter is applied. This filter computes a texture score $s_t$ for each image pixel, which reflects the texture intensity within a local neighborhood. Pixels for which this score is below a configurable threshold $t_t$ are again labeled as invalid in the computed disparity map.

\subsubsection{\small Speckle Filtering}
\label{sec:speckle_filt}

The aforementioned methods are not always able to identify and label all erroneous matches. Fortunately, the erroneous matches that remain tend to appear as small clusters of similar disparity. These \textit{speckles} are then removed with a speckle filter. The speckle filter identifies connected components that are below a specified minimum size. The minimum speckle size is controlled through the speckle filter window size $w_s$. The pixels that belong to identified speckles are again labeled as invalid.

\subsubsection{\small Gap Interpolation}
\label{sec:gap_interpol}

The aforementioned post-processing techniques all remove pixels form the computed disparity map, which leaves gaps with no valid disparity data. If one such gap is small, it can be filled with valid disparities by interpolating the disparities from its edges. Interpolation is only performed for gaps whose vertical and horizontal extent $l_h$ and $l_v$ fulfill the condition
\begin{equation}
\min \left\{ l_h, l_v \right\} \le l_{max} \mbox{,}
\end{equation}
where $l_{max}$ is the maximum gap width. Interpolation is also omitted if the disparities from the edge of the identified gap do not have a similar magnitude.

\subsubsection{\small Noise Reduction}
\label{sec:median}

Finally, a noise reduction filter is applied to the generated disparity map. This filter performs a smoothing of the disparity map, while being aware of discontinuities and invalid disparities. The processing result of this filter is directly transmitted as output through SceneScan's ethernet port.

\section{Results}

\begin{table}[p]
	\centering
	\begin{tabular}{r|rrrr}
	\multicolumn{1}{l|}{\textbf{Disparity~}} & \multicolumn{4}{c}{\textbf{Image Resolution}}\\
    \multicolumn{1}{l|}{\textbf{Range}} & ~~~640 $\times$ 480 & ~~~800 $\times$ 592 & ~~~1280 $\times$ 960 & ~~~1600 $\times$ 1200\\
	\hline
    128 pixels~ & 100 fps & 65 fps & 24 fps & 15 fps\\
    256 pixels~ & 70 fps & 45 fps & 15 fps & 10 fps\\
    \end{tabular}
	\caption{SceneScan Pro processing performance.}
	\label{tab:scenescan_fps}
\end{table}

\begin{figure}[p]
    \centering
    \subfigure[]{
        \!\!\!
        \includegraphics[height=3cm]{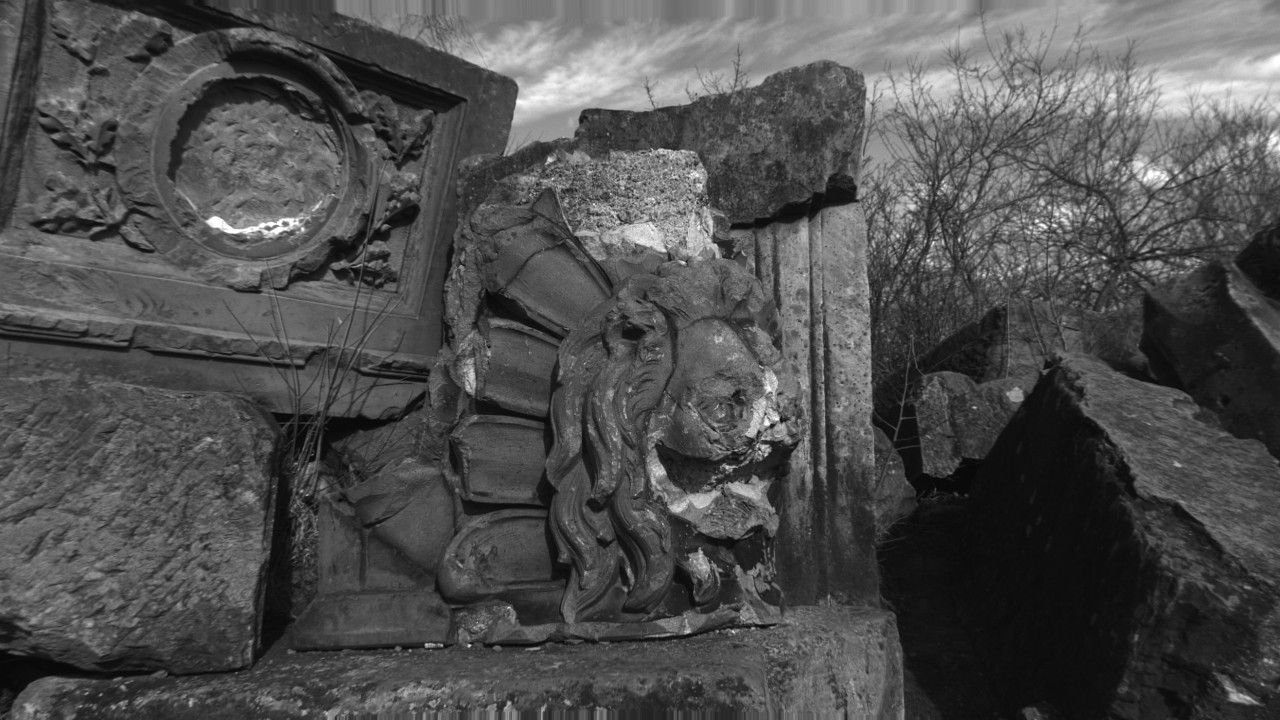}
        \!\!\!
    }
    \subfigure[]{
        \!\!\!
        \includegraphics[height=3cm]{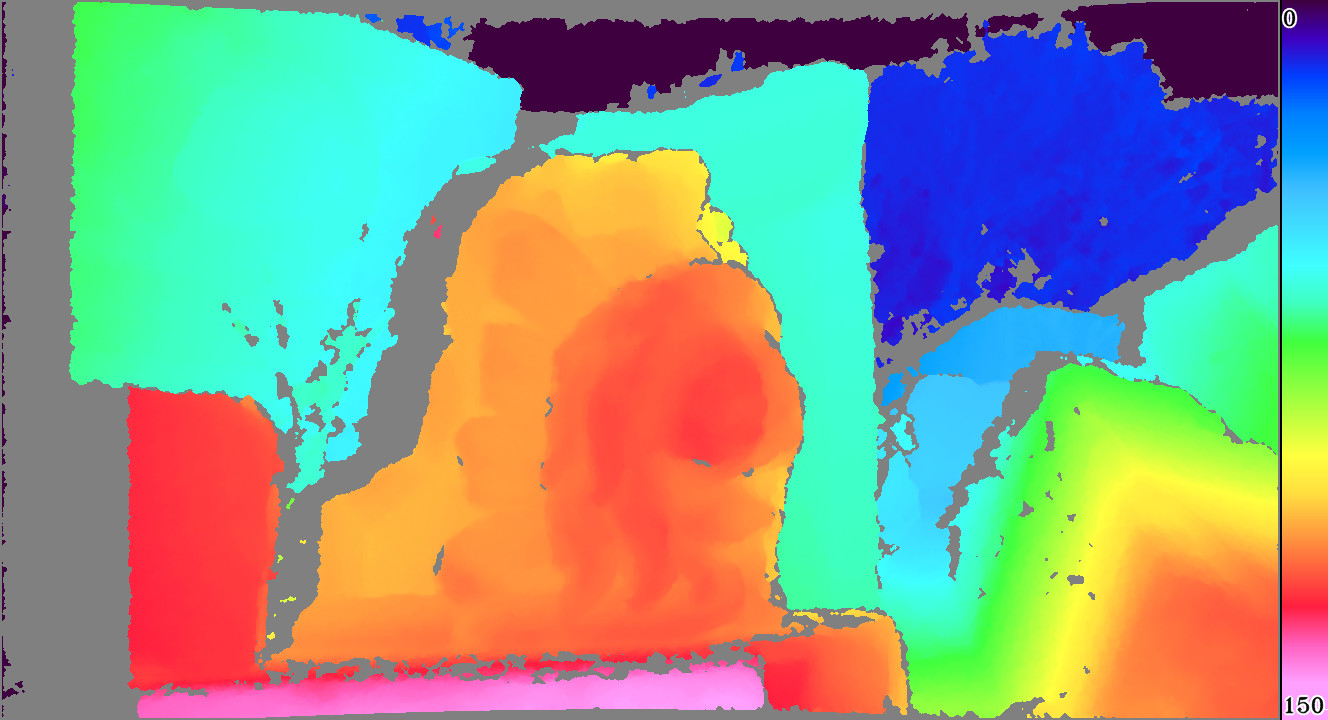}
        \!\!\!
    }
    \subfigure[]{
        \!\!\!
        \includegraphics[height=3cm]{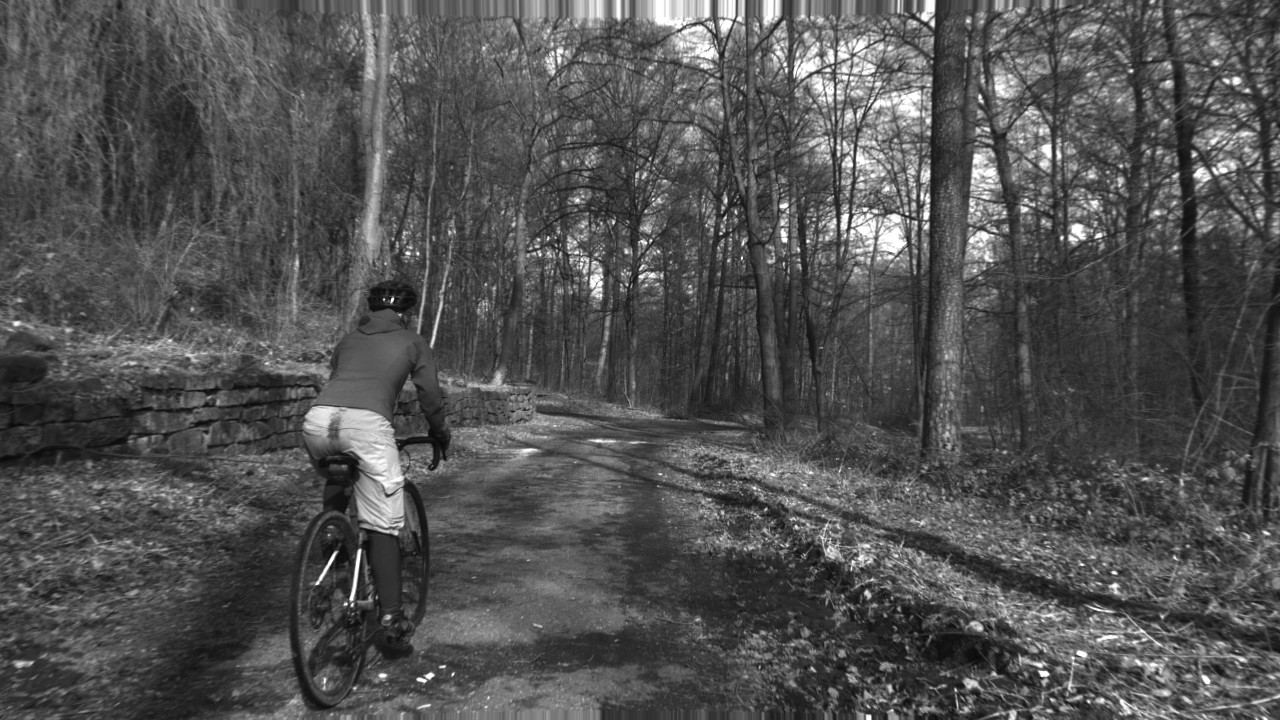}
        \!\!\!
    }
    \subfigure[]{
        \!\!\!
        \includegraphics[height=3cm]{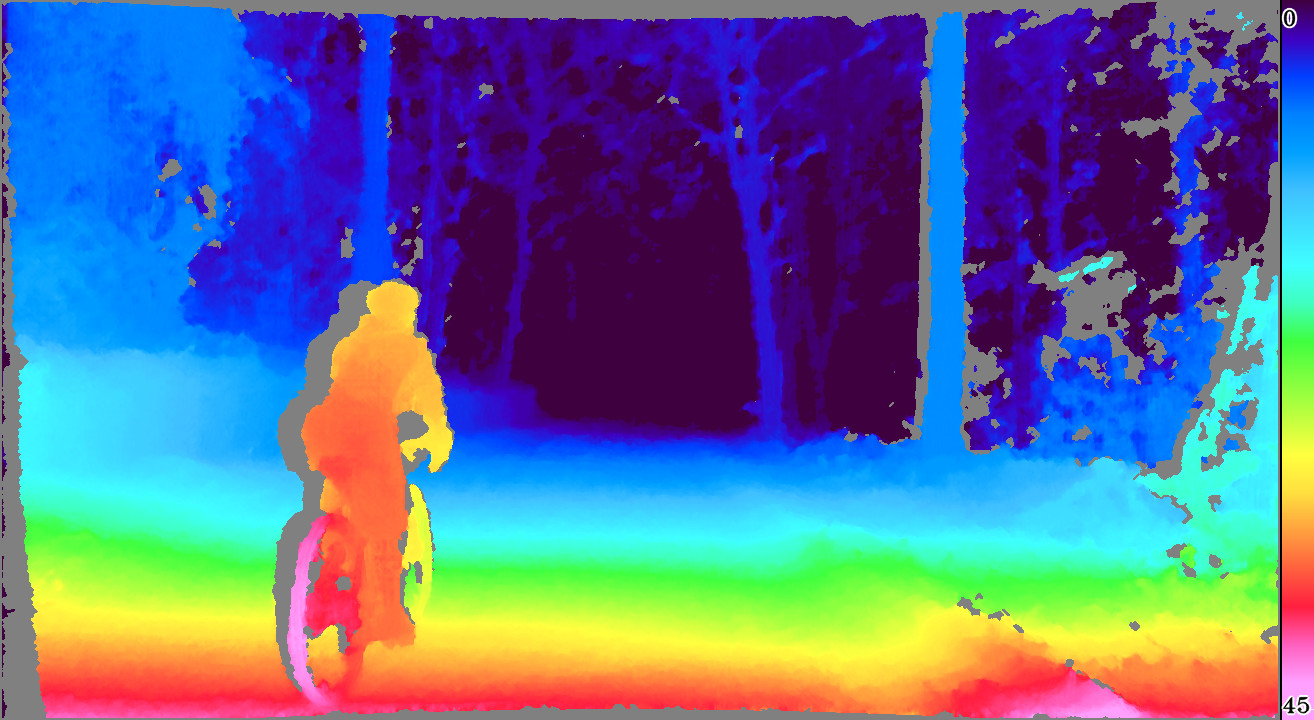}
        \!\!\!
        \label{fig:dispmap2}
    }
    \subfigure[]{
        \!\!\!
        \includegraphics[height=3cm]{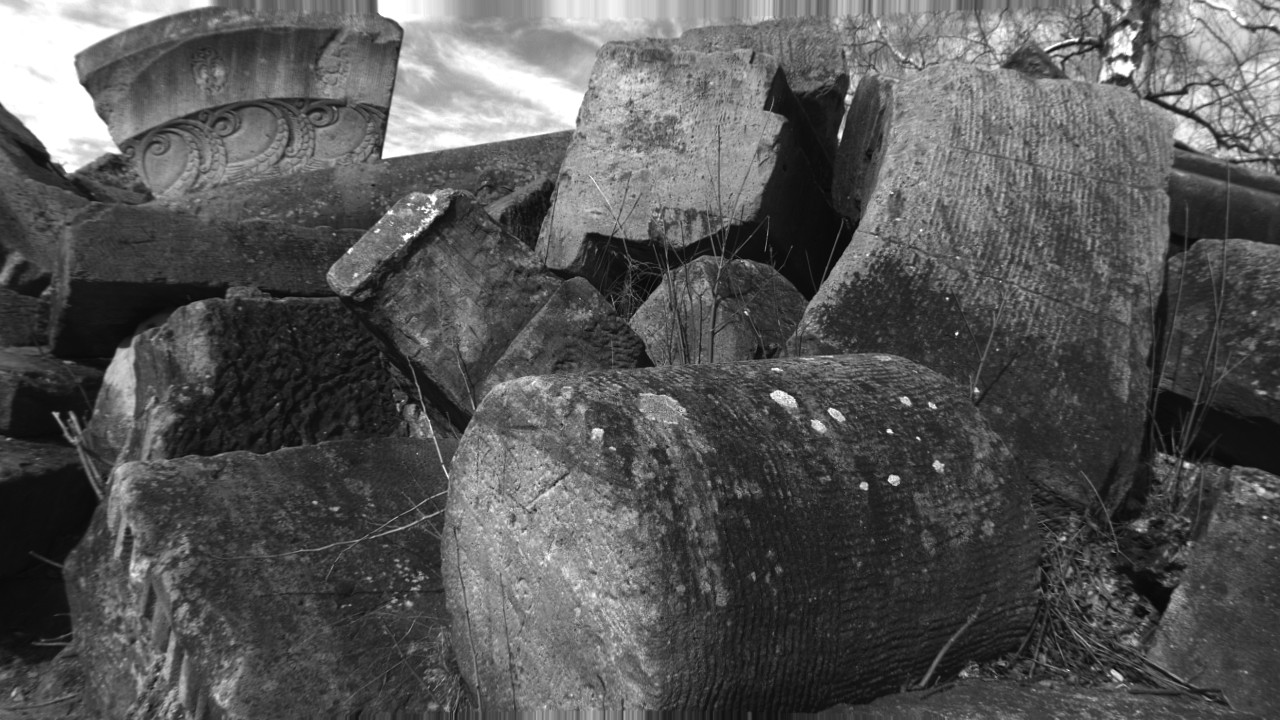}
        \!\!\!
    }
    \subfigure[]{
        \!\!\!
        \includegraphics[height=3cm]{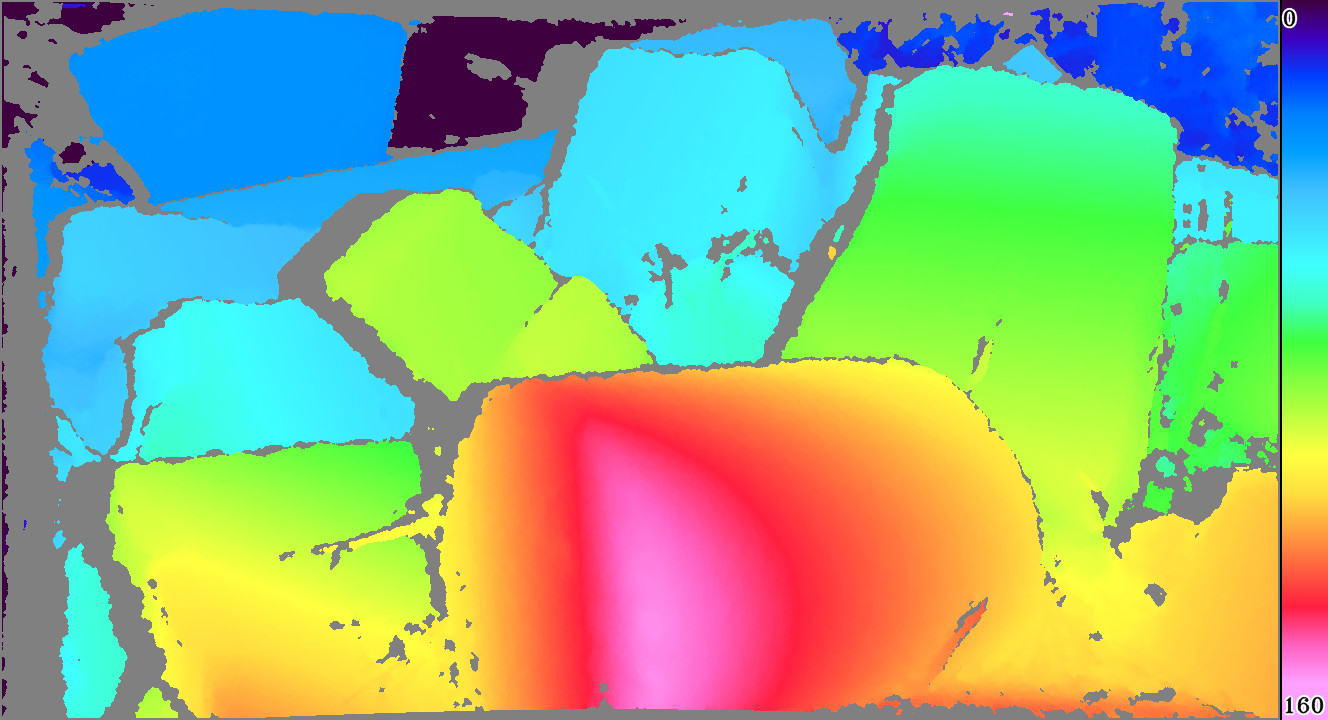}
        \!\!\!
    }
    \caption {(a,c,e) Examples for left input images and (b,d,f) computed disparity maps.}
    \label{fig:examples}
\end{figure}

Table~\ref{tab:scenescan_fps} lists the achievable frame rates of SceneScan Pro on different image resolutions and disparity ranges. These numbers represent the frame rates at which a connected computer would receive the disparity data by SceneScan Pro, and hence already include all overheads for camera and network communication. At all configurations, SceneScan Pro requires only 8 W of power, without supplying power to the cameras.

The highest frame rate can be achieved at an image resolution of 640~$\times$~480 with a 128 pixels disparity range. In this configuration SceneScan Pro can provide up to 100 fps. With increasing resolution and disparity range, the frame rate decreases. The maximum image resolution which SceneScan Pro can process is 1856 $\times$ 1856, which is 3.4 megapixels.

When using the full 256 pixels disparity range, SceneScan Pro is able to perform 5.1 billion disparity evaluations per second, and it achieves an output of 20 million disparities per second. If the disparity range is reduced to 128 pixels, the processing performance increases to 30 million disparities per second. As each disparity value can be projected to a 3D coordinate, this means that SceneScan Pro provides 30 million 3D point measurements per second.

In Figure~\ref{fig:examples} we can see a collection of example input images and the corresponding disparity maps that were computed with SceneScan Pro. All images were recorded outdoors on a bright day. As stereo vision does not struggle with bright ambient light, we do receive very dense disparity maps for all examples.

Each example image contains sections with very close objects, as well as sections with distant objects. As can be seen, the distance of an object does not influence the measurement density or robustness. Even for very distant image regions, such as in the center of Figure~\ref{fig:dispmap2} we do receive very dense disparity measurements.

Overall the disparity maps appear very smooth with only very few outliers. This can be credited to the extensive post-processing of the cost cube and the disparity maps. Occluded image regions are  effectively filtered, as can be seen by the clear occlusion shadow to the left of all foreground objects in Figure~\ref{fig:examples}.

\section{Conclusion}

With SceneScan and SceneScan Pro we have created two stand-alone systems for stereo vision that use FPGAs as the underlying computation hardware. By using FGPAs, our systems are able to achieve a very high-power efficiency. At the same time, they provide an exceptional computation performance, and are thus able to process camera images at frame rates up to 100 fps.

Particularly in the field of autonomous mobile robots, accurate high-speed depth estimates are a prerequisite for many applications. Stereo vision promises to provide such data even in bright environments and over long measurement distances.

Our SceneScan and SceneScan Pro systems can easily be integrated into battery-powered and energy constrained mobile platforms, where large power-hungry GPU-based systems are not an option. We hope that by introducing SceneScan and SceneScan Pro, we have made stereo vision more accessible to a wide range of researchers and developers. This will hopefully facilitate the development of more systems that make use of the principles of stereo vision.

\bibliographystyle{IEEEtran}    
\bibliography{schauwecker}{}

\cleardoublepage